\documentclass{article}
\usepackage{ijcai24}

\usepackage{times}
\usepackage{soul}
\usepackage{url}
\usepackage[hidelinks]{hyperref}
\usepackage[utf8]{inputenc}
\usepackage[small]{caption}
\usepackage{graphicx}
\usepackage{amsmath}
\usepackage{amsthm}
\usepackage{booktabs}
\usepackage{algorithm}
\usepackage{algorithmic}
\usepackage[switch]{lineno}

\usepackage{multirow}
\usepackage{subfigure}
\usepackage{adjustbox}
\usepackage{microtype}
\usepackage{inconsolata}
\usepackage{newtxtext,newtxmath}
\usepackage{bbold}
\usepackage{mathtools}
\usepackage{bm}
\usepackage{makecell}
\usepackage{latexsym}
\usepackage[dvipsnames]{xcolor}

\newcommand{\wikirecent}{Recent}
\newcommand{\wikicf}{CounterFact}
\newcommand{\zsre}{ZsRE}
\newcommand{\convsent}{ConvSent}

\usepackage{graphicx,calc}
\usepackage{wrapfig}

\newlength\myheight
\newlength\mydepth
\settototalheight\myheight{Xygp}
\title{\textsc{InstructEdit}: Instruction-Based Knowledge Editing for Large Language Models}

\author{
\textbf{Ningyu Zhang}$^{\clubsuit}$\footnotemark[1], Bozhong Tian$^{\clubsuit}$\thanks{~~Equal contribution.}, Siyuan Cheng$^{\heartsuit}$\footnotemark[1], Xiaozhuan Liang$^{\heartsuit}$\footnotemark[1], \\
Yi Hu$^{\heartsuit}$,
Kouying Xue$^{\heartsuit}$, Yanjie Gou$^{\heartsuit}$, Xi Chen$^{\heartsuit}$\thanks{~~Corresponding author.}, 
{\bf Huajun Chen}$^{\clubsuit
}$\footnotemark[2], 
\affiliations
 $^\clubsuit$ Zhejiang University
$^\heartsuit$ Tencent
\emails
\{zhangningyu, tbozhong\}@zju.edu.cn\\
  \textcolor{blue}{\url{https://zjunlp.github.io/project/InstructEdit}}
}

\begin{document}

\maketitle

\begin{abstract}
Knowledge editing for large language models can offer an efficient solution to alter a model’s behavior without negatively impacting the overall performance. However, the current approaches encounter issues with limited generalizability across tasks, necessitating \textbf{one distinct editor for each task}, significantly hindering the broader applications. To address this, we take the first step to analyze the multi-task generalization issue in knowledge editing. Specifically, we develop an instruction-based editing technique, termed \textsc{InstructEdit}, which facilitates the editor's adaptation to various task performances simultaneously using simple instructions. With only one unified editor for each LLM, we empirically demonstrate that \textsc{InstructEdit} can improve the editor's control, leading to an average 14.86\% increase in Reliability in multi-task editing setting. Furthermore, experiments involving holdout unseen task illustrate that \textsc{InstructEdit} consistently surpass previous strong baselines. To further investigate the underlying mechanisms of instruction-based knowledge editing, we analyze the principal components of the editing gradient directions, which unveils that instructions can help control optimization direction with stronger OOD generalization\footnote{~~Code and datasets are available in \url{https://github.com/zjunlp/EasyEdit}.}.
\end{abstract}

\section{Introduction}

\begin{figure}[t]
    \centering
    \includegraphics[width=0.49\textwidth]{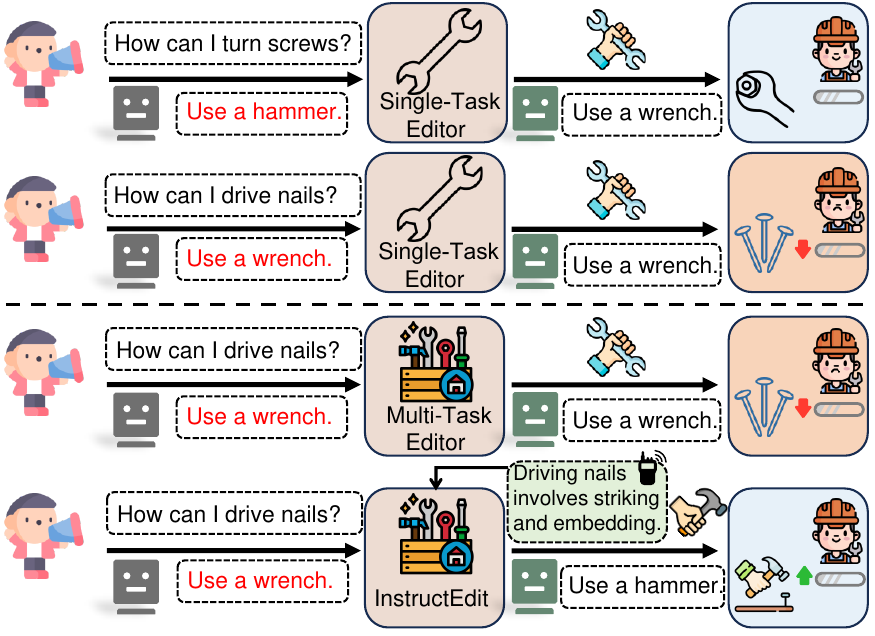}
    \caption{
\textbf{Top:} The Single-Task Editor excels in specific tasks (e.g., turning screws) but fails in others (e.g., driving nails). 
 \textbf{Bottom:} The vanilla Multi-Task Editor (all data mixed together) still struggles to choose the right tool for varied tasks without aid. 
    Thus, we propose \textsc{InstructEdit}, enabling the Multi-Task Editor to respond aptly (such as using a hammer for nails) with instructional guidance.}
    \label{fig:intro}
\end{figure}

Knowledge editing~\cite{Editable_Neural_Network,yao2023editing,wang2023knowledge,mazzia2023survey,si2023knowledge,zhang-etal-2023-editing,zhang2024comprehensive} aims to enable efficient and targeted post-hoc modifications in the parametric knowledge within Large Language Models (LLMs)~\cite{MEND,KnowledgeNeurons,GRACE,MMEdit,tan2023massive}. 
For example, as shown in Figure \ref{fig:intro}, when prompting with ``How can I turn screws?'', knowledge editing techniques can focus on specific areas in LLMs for adjustment, changing the answer from ``Use a hammer'' to ``Use a wrench''  without compromising the overall performance.
Recently, numerous works on knowledge editing for LLMs have been proposed, which can be divided into two main paradigms \cite{yao2023editing}: 1) Preserve Models’ Parameters by utilizing additional parameters or memory \cite{SERAC}; 2) Modify Models’ Parameters to alter the weights responsible for the undesirable output \cite{ROME}.

However, previous knowledge editing approaches mainly focus on \textbf{single-task settings}, which means they may fail to achieve \textbf{multi-task generalization} capabilities and demonstrate inefficiency in editing when confronted with Out-of-Distribution (OOD) data.
For example, as shown in Figure \ref{fig:intro} and Table \ref{tab:ood_results}, the knowledge editing approach can simply change the behavior when prompting with ``How can I turn screws'', but fail to generalize to different task when prompting with ``How can I drive nails''.
Fundamentally, for the Preserve Models' Parameters paradigm, Additional Parameters methods~\cite{CALINET,LoRA,T-patcher} fit updated data with few extra parameters, while Memory-based approaches~\cite{SERAC,GRACE}, storing only current batch knowledge, can hardly generalize to OOD data.
For the Modify Models' Parameters paradigm, Locate-Then-Edit~\cite{ROME,MEMIT} target and directly update specific parameters, but their updates are confined to provided data, limiting the model's generalization to other domains.
Meta-learning editing approaches~\cite{KnowledgeEditor,MEND,cheng2023editing_} represent a branch in the realm of the Modify Models’ Parameters paradigm, which utilizes a hypernet to predict specific weight updates for each data point, thereby facilitating the editing of LLMs~\cite{GPT-2,LLMs_survey,LLaMA}.
Yet traditional meta-learning editing methods typically focus on training a hypernet, which in essence functions as the Editor, specialized for a particular domain. 
Consequently, knowledge editing for a new task demands re-training the Editor, resulting in significant computational costs.

\begin{table}
\centering
\resizebox{0.97\columnwidth}{!}{
\begin{tabular}{ccccc}
\toprule
Unseen & Seen & Reliability  & Generalization  &  Portability  \\
\cmidrule{1-5}
\multirow{3}{*}{\textbf{CounterFact}} &  \textbf{CounterFact} &  84.62 & 46.01 & 42.46\\
 &  \textbf{Recent} &  \raisebox{-\mydepth}{\includegraphics[height=1.0\myheight]{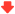}} -25.50  & \raisebox{-\mydepth}{\includegraphics[height=1.0\myheight]{figs/downarrow.png}} -21.34  
 & \raisebox{-\mydepth}{\includegraphics[height=1.0\myheight]{figs/downarrow.png}} -7.33 \\
 &  \textbf{ZsRE}    &  \raisebox{-\mydepth}{\includegraphics[height=1.0\myheight]{figs/downarrow.png}} -25.26  & \raisebox{-\mydepth}{\includegraphics[height=1.0\myheight]{figs/downarrow.png}} -18.36  
 & \raisebox{-\mydepth}{\includegraphics[height=1.0\myheight]{figs/downarrow.png}} -4.79 \\
\cmidrule{1-5}
\multirow{3}{*}{\textbf{ZsRE}} & \textbf{ZsRE} & 96.62 & 94.60 & 48.85 \\
 & \textbf{Recent}    &  \raisebox{-\mydepth}{\includegraphics[height=1.0\myheight]{figs/downarrow.png}} -86.40  & \raisebox{-\mydepth}{\includegraphics[height=1.0\myheight]{figs/downarrow.png}} -91.33  
 & \raisebox{-\mydepth}{\includegraphics[height=1.0\myheight]{figs/downarrow.png}} -0.60 \\
 & \textbf{CounterFact}    &  \raisebox{-\mydepth}{\includegraphics[height=1.0\myheight]{figs/downarrow.png}} -56.31 
 & \raisebox{-\mydepth}{\includegraphics[height=1.0\myheight]{figs/downarrow.png}} -64.90 
 & \raisebox{-\mydepth}{\includegraphics[height=1.0\myheight]{figs/downarrow.png}} -1.35  \\
\bottomrule
\end{tabular}
}
\caption{
Motivating knowledge editing results in multi-task generalization. 
Directly transferring to the unseen task (CounterFact and ZsRE) can result in a significant performance decay.}
\label{tab:ood_results}
\end{table}

Intuitively, devising a strategy to enable the knowledge editing methods to effectively generalize across tasks is beneficial.
Reflecting on prior research, to enhance the model's generalization capabilities, researchers have introduced instruction tuning~\cite{FLAN}. 
Instruction tuning can enhance the LLMs' comprehension skills by providing clearer commands or instructions, enabling the model to understand better and execute accurate responses.
Previous studies \cite{FLAN,instruction_tuning_survey} observe that models refined through instruction tuning not only excel in performance on in-distribution datasets but also effectively generalize to previously unseen instruction data.
Inspired by this, we propose the \textbf{Instruct}ion-based \textbf{Edit}ing method, dubbed as \textsc{InstructEdit}, which learns a well-formed Editor by designing the corresponding instructions for training on different tasks\footnote{The instructions text in this paper are limited to task descriptions rather than natural language instructions, which is a limitation we leave for future work.}
, as shown in Figure \ref{fig:intro}.
Specifically, we utilize meta-learning editing methods to train the editor across various meticulously curated instructions.
We conduct experiments on four datasets and observe that \textsc{InstructEdit} can equip the Editor with the capability for multi-tasking editing, thereby conserving substantial human and computational resources. 
Our experiments reveal that \textsc{InstructEdit} can enhance the reliability by 14.86\% (compared with MEND) on average when editing GPT2-XL. 
Furthermore, it can yield improvement by 42.04\% on OOD dataset unseen during training.

\section{Related Work}
\subsection{Knowledge Editing}

Recently, knowledge editing~\cite{Editable_Neural_Network,zhang2024comprehensive} has emerged, aiming for efficient and accurate updates of knowledge in LLMs, to address the issues of outdated knowledge due to their training cut-off, factual fallacy, and potential generation of unsafe content.
This technique is applied in various domains \cite{Xu2022LanguageAC,Edit_Personality,hase2023does,wang2023crosslingual,Li2023PMETPM,MMEdit,zhong2023mquake,akyürek2023dune,si2024mpn}, with an increasing number of researches investigating the impact of knowledge editing~\cite{ilharco2023editing,DBLP:journals/corr/abs-2305-14956,DBLP:journals/corr/abs-2307-12976,DBLP:journals/corr/abs-2308-09954,easyedit,erasing_concept,brown2023edit,wei2023assessing,pan2023finding,li2023evaluating,li2023inferencetime,ju2023klob,li2023unveiling,onoe2023lms,pinter2023emptying,gupta2024model,hernandez2023linearity,huang2024unseen,gu2024model,lo2024large,yin2023history,DBLP:journals/corr/abs-2312-11795,ma2024neighboring}.
Researchers have diligently classified existing knowledge editing approaches into two main paradigms:

\paragraph{Preserve Models’ Parameters.} 
For those approaches, knowledge can be updated without altering models' parameters, primarily following two paradigms: \texttt{Additional Parameters} and \texttt{Memory Based}.
\texttt{Additional Parameters} integrate extra trainable parameters into the models. 
These added parameters are trained on a modified knowledge dataset, while the original models parameters remain unchanged.
T-Patcher~\cite{T-patcher} embeds a single neuron (patch) for each error in the model's final Feed-Forward Network (FFN) layer, activating only upon encountering the respective mistake.
CaliNet~\cite{CALINET} drawing inspiration from \cite{KnowledgeNeurons}, 
introduces additional trainable parameters into the FFNs.
\texttt{Memory Based} store edit examples in memory and use a retriever to select relevant edit facts for new inputs, thereby directing the model's fact generation.
SERAC~\cite{SERAC} presents a method that utilizes a distinct \emph{counterfactual model} while maintaining the integrity of the original model.
GRACE~\cite{GRACE} employs a distinct codebook as an adapter, progressively incorporating and refreshing elements to refine the model's predictions.
In-context Knowledge Editing~\cite{IKE} produces outputs aligned with given knowledge using refined in-context prompts.  

\paragraph{Modify Models’ Parameters.}
Those approaches edit LLMs by modifying a portion of the parameter $\theta$ via applying an $\Delta$ matrix.
There are primarily two paradigms: \texttt{Locate-Then-Edit} and \texttt{Meta-learning}.
\texttt{Locate-Then-Edit} targets and directly updates specific parameters. 
ROME~\cite{ROME} utilizes causal mediation analysis for targeted editing but is limited to one fact at a time. 
Addressing this, MEMIT~\cite{MEMIT} has been proposed, an advancement of ROME, enabling direct memory embedding into the model through rank-one modifications of single-layer MLP weights.
\texttt{Meta-learning} utilizes a hypernet to predict specific weight updates for each data point.
MEND~\cite{MEND} and Knowledge Editor (KE)~\cite{KnowledgeEditor} 
propose strategies that include an external editor, adept at identifying the optimal parameter set, $\theta$, for knowledge editing, whilst simultaneously enforcing constraints to preserve the stability of the model.

\subsection{Instruction Tuning}
Instruction Tuning~\cite{instruction_tuning_survey} markedly improves models' capability to handle new and unseen tasks by teaching them to comprehend and follow natural language instructions.
In NLP, the focus is rapidly shifting towards refining LLMs~\cite{GPT-3,chatgpt,sun2023moss,alpaca,su2023embedder} to follow natural language instructions for real-world tasks.
The effectiveness of these approaches is evident in the enhanced zero-shot and few-shot learning capabilities of these LLMs, demonstrating their improved proficiency in adapting to new tasks with minimal prior exposure.
\begin{table}[!t]
\centering
\resizebox{0.48\textwidth}{!}{
\begin{tabular}{ll}
\toprule
\textbf{Task (Dataset)}  & \textbf{Instruction}      \\
\midrule
\multirow{4}{*}{\textbf{\wikicf}}  
& {\color{RoyalBlue}\textbf{Task}}: CounterFact
\\& {\color{ForestGreen}\textbf{Description}}: A dataset designed to
\\& challenge and assess model on...
\\&  {\color{Fuchsia}\textbf{Input}}: The official language of...
\\

\midrule
\multirow{4}{*}{\textbf{\convsent}} 
& {\color{RoyalBlue}\textbf{Task}}: ConvSent
\\&    {\color{ForestGreen}\textbf{Description}}: Teach the chatbot to
\\&   sound [LABEL] about [TOPIC]...
\\&   {\color{Fuchsia}\textbf{Input}}: What do you think of...
\\

\midrule
\multirow{1}{*}{\textbf{...}}  
& ...
\\

\bottomrule
\end{tabular}
}
\caption{Examples of the instructions.
As for \text{\convsent}, we need to replace [LABEL] and [TOPIC] according to the input.} 
\label{table:instruct_prompt}
\end{table}
Inspired by the generalization capabilities of Instruction Tuning~\cite{liang2023contrastive,instructgpt,instruction_tuning_survey}, we take the first step to integrate instructions into knowledge editing for LLMs, endowing one unified Editor with commendable instruction generalization and zero-shot capabilities to concurrently handle multiple editing tasks.

\section{Background}
\paragraph{Knowledge Editing Task Definition.}
Knowledge editing, as described by~\cite{zhang2024comprehensive}, aims to alter the behavior of an initial base model $f_{\theta}$ (where $\theta$ represents the model's parameters) in reaction to a specific edit descriptor $(x_i, y_i)$ while maintaining the model's performance on other samples.
The target is to create an edited model $f_{\theta'}$,
which succinctly encapsulates the intended modifications in the model's performance.
Concretely, the model $f_{\theta}$ can be represented with a function $f: \mathbb{X} \rightarrow \mathbb{Y}$ which associates an input $x$ with its corresponding prediction $y$. 
Given an edit descriptor that includes the edit input $x_i$ and edit label $y_i$ with the condition that $f_{\theta}(x_i) \neq y_i$, the revised model 
$f_{\theta'}$ is engineered to yield the anticipated output, ensuring that $f_{\theta'}(x_i) = y_i$.
 
 
\paragraph{Multi-Task Editing Definition.}
In this paper, we mainly focus on multi-task editing setting, which means the editing approach should have the ability to handle various multiple tasks.
We denote a LLM as $f$, characterized by its parameters $\theta$ to form $f_{\theta}$. 
For editing in a single task, we introduce a dataset as $D_{edit}$. 
When we extend to multi-tasking scenarios, the dataset becomes a set comprising a collection $\{D_{edit}^{t_1}, D_{edit}^{t_2}, ..., D_{edit}^{t_j}\} \sim \mathcal{T}$, with each element representing to a unique task. 
In each specific task $t_j$, we engage with original input-output knowledge pairs, expressed as $(x_i^{t_j}, y_i^{t_j}) \sim D_{edit}^{t_j}$. 
The editing objective is to evolve the model's output from the original erroneous $y_i^{\prime}$ to a more accurate $y_i^{t_j}$, achieved by adjusting the model's parameters from $f_{\theta}$ to $f_{\theta'}$. 
Formally, the procedure can be described as follows:
\begin{equation}
 f_{\theta}(x_i^{t_j})=y_i^{\prime} \rightarrow f_{\theta'}(x_i^{t_j})=y_i^{t_j}   
\end{equation}

Note that for all experiments, we utilize the multi-task editing setting and report the performance in Table \ref{tab:main-results}.
We also select one unseen dataset (a.k.a., ZsRE is unseen when training the Editor) for hold out editing evaluation.

\begin{figure*}[t]
\centering 
\includegraphics[width=0.99\textwidth]{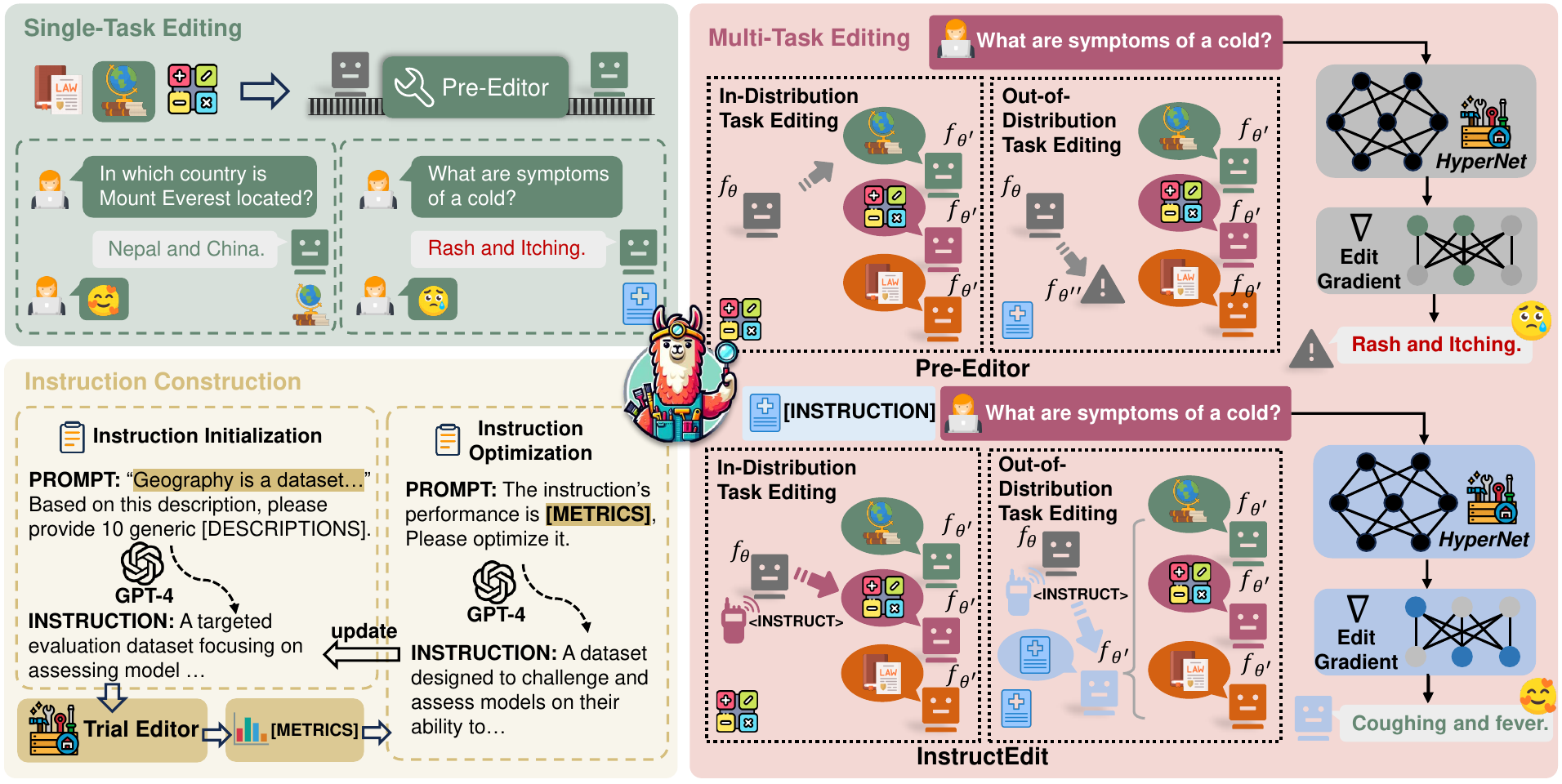}
\caption{
Assuming access to multi-domain task data: Law, Geography, Medicine, and Math.
\textbf{Single-Task Editing}) Original editing is domain-specific (e.g., a Geography Editor edits geography-related knowledge but can't transfer it to Medicine).
\textbf{Multi-Task Editing}) Previous methods (Pre-Editor) trained across domains (Law, Geography, and Math) often misdirect In-Distribution Task Editing. For OOD Task Editing (Medicine), a lack of guidance $\nabla$ leads to missing the correct edit region. Instructions enable precise editing and improve generalization.
\textbf{Instruction Construction}) We utilize GPT-4 to generate instructions through well-crafted prompts, evaluate metrics using the Trial Editor, and then employ GPT-4 for continuous Instruction Optimization, enhancing the instructions until there is no further improvement in metrics.
}
\label{fig:multitask_editing}
\end{figure*}

\section{Instruction-Based Knowledge Editing}
\label{sec:instructedit_all}

\subsection{Instruction Dataset Construction}
\paragraph{Selected Task.}
To ensure diversity in multi-task editing, we select a range of datasets: \text{\wikirecent}~\cite{zhang2024comprehensive} for knowledge insertion, \text{\wikicf}~\cite{zhang2024comprehensive} for counterfactual generation, and \text{\convsent}~\cite{SERAC} for sentiment editing in knowledge updating.

\textbf{\wikirecent} focusing on triplets added to \textsc{WikiData} after July 2022, is used to enable model updates with the latest knowledge. 

\textbf{\wikicf} emphasizes triplets from top-viewed Wikipedia pages to address the issue of models overlooking less prominent entities in modification edits.

\textbf{\convsent} is a sentiment editing task aimed at adjusting a dialog agent's sentiment on a specific topic, like ``What do you think of bananas?'' without affecting responses of other topics. The training approach retains the original settings of the \text{\convsent}.
Additionally, we utilize a balanced subset, randomly sampled from the original \convsent, for multi-task training. Detailed analyses are presented in Figure \ref{fig:task_num}. 

\paragraph{Hold Out Task.}
Empirically, we find that transferring knowledge from other tasks to \text{\zsre} is challenging as shown in Table \ref{tab:ood_results}.
Therefore, we utilize \zsre, a zero-shot relation extraction dataset, to evaluate the generalization ability of multi-task editing, which means we do not incorporate \zsre in multi-task editing training.
Specifically, we use the extended version by~\cite{yao2023editing}, which adds a portability test and new locality sets to the original dataset.

\paragraph{Instruction Generation.}
We develop instruction templates for multi-task knowledge editing, encompassing four task families, they are: \wikicf, \wikirecent, \convsent, and \zsre.
Each includes instructions for task-specific model discovery, with input and target templates, and associated metadata.
Specifically, we craft tailored instruction sets for each task family, including \texttt{[Task]}, \texttt{[Description]}, and \texttt{[Input]}.
The \texttt{[Task]} represents the specific task linked to a data item, while the \texttt{[Input]} embodies the data item itself.
We delve into the specifics with the \texttt{[Description]}, which is the essential component that uniquely tailors each task.
Leveraging GPT-4 and detailed task information, we generate 20 descriptions for each task and manually select 10 candidates based on their clarity and conciseness. 
Subsequently, we concatenate \texttt{[Task]}, \texttt{[Description]}, and \texttt{[Input]} to form the instructions presented in Table \ref{table:instruct_prompt}.
Notably, while the last instruction is used to evaluate the model's generalization capabilities with instructions, the others are utilized for training.
We further optimize instructions by feeding them with performance metrics into GPT-4 to improve the quality as shown in Figure \ref{fig:multitask_editing}.  
All instruction data will be released to the community.

\subsection{Unified Editor Learning with Instructions}
\label{sec:instructedit}

In this section, we primarily focus on the crucial role of instructions in directing the editing process and delve into a detailed explanation of how \textsc{InstructEdit} works.
Specifically, we define the instruction set as $\{I^{t_1}, I^{t_2}, ..., I^{t_j}\} \sim \mathcal{I}$, where $I^{t_j}$ represents a collection of instructions for task $t_j$. 
Based on the instructions, we outline the editing process as follows:
\begin{equation}
\begin{cases}
f_{\theta'}(in^{t_j}, x_i)=y_i^{t_j} & x_i \in E(x_i^{t_j}), x_i^{t_j} \in  D_{edit}^{t_j}\\
f_{\theta'}(x_i)=f_{\theta}(x_i) & Otherwise
\end{cases}
\end{equation}
where $in^{t_j}$ refers to an instruction randomly selected from $I^{t_j}$, $E(x_i^{t_j})$ includes both $x_i^{t_j}$ and its equivalent expressions.
\textsc{InstructEdit} employs the editing architecture of MEND, utilizing a meta-learning editor (hypernetwork) for implementing edits. 
\textsc{InstructEdit} updates the model's parameters $u_\ell \in \mathcal{M}$ with an editor parameterized by $\phi$. 
It does this by mapping $u_\ell^i$ (the input to layer $\ell$ for each batch element $i$) and the gradient $\delta_{\ell+1}^i$ (calculated as $\delta_{\ell+1}^i \gets \nabla_{W_\ell} L(x_i, y_i)$) to \textit{pseudoactivations} $\tilde u_\ell^i$ and \textit{pseudodelta} $\tilde\delta_{\ell+1}^i$.
The knowledge editing gradient for the weight matrix $u_\ell$ is then represented as follows:
\begin{equation}
    \label{eq:pseudogradient}
    \textstyle \tilde{\nabla}_{u_\ell} = \tilde\delta_{\ell + 1}^{i}{\tilde u_\ell^{i\,\top}}.
\end{equation}

Additionally, we scale the gradient $\tilde{\nabla}_{u_\ell}$ with $L_2$ norm of the gradient to isolate its directional component, denoted by $\vec{\nabla}_{u_\ell} = \tilde{\nabla}_{u_\ell} \, / \, \|\tilde{\nabla}_{u_\ell}\|_2$. 
Intuitively, $\vec{\nabla}_{u_\ell}$ pinpoints the key knowledge area for editing elements $i$.
This facilitates a more meaningful comparison across various tasks by focusing solely on the gradient's direction while discarding its magnitude.
We term this focused area as \textbf{editing area}.

Our primary objective is to equip the editor with the ability to comprehend and apply editing instructions, thus enhancing its capability to edit tasks that fall outside the usual distribution.
Additionally, we append instructions before the input to facilitate multi-task editing.
\textsc{InstructEdit} aims to augment multi-task editing capabilities, seeking a synergistic impact where the collective result surpasses the individual contributions.
Through the concatenation of instructions, as shown in Figure \ref{fig:multitask_editing}, \textsc{InstructEdit} aims to cluster task vectors and reduce conflicts between tasks, which guarantees that the performance of the multi-task editor on individual tasks matches or even surpasses that of dedicated single-task editors.


\begin{table*}[t]
{
\centering
\setlength{\tabcolsep}{2.0mm}{
\begin{adjustbox}{width=0.85\textwidth,center}
\begin{tabular}{lcc|c|ccccc|c}
\toprule
\textbf{DataSet}     & \textbf{Model}                                      & \textbf{Metric}   &  \textbf{Base} 
& $\textbf{FT-L}$ & $\textbf{CaliNet}$ & $\textbf{KE}$
& $\textbf{MEND}$ & $\textbf{GRACE}$   & $\textbf{\textsc{InstructEdit}}$  \\ 
\midrule
\multicolumn{10}{c}{\textbf{\small Multi-Task Editing}} \\
\midrule
\multirow{7}{*}{\textbf{\wikicf}}        & \multirow{3}{*}{GPT2-XL}                  
& Reliability   & 0.00 & 0.40 & 0.24 & 33.97 & 74.26 & \textbf{96.31} & \underline{80.81}  \\
&

& Generalization   & 0.00 & 0.32 & 0.12 & 8.70 & \underline{46.48} & 0.00 & \textbf{53.16}  \\
                             &                                           
& Locality   & 100.0 & 43.73 & 82.81 & \underline{90.94} & 58.68 & \textbf{99.99} & 67.83  \\
                             &                                           
& Portability   & 11.00 & 0.87 & 3.64 & 27.41 & \underline{41.88} & 11.00 & \textbf{50.83} \\ 
\cmidrule{2-10} 
                             & \multirow{3}{*}{LLaMA-2}                     
& Reliability   & 0.00 & 0.00 & 0.00 & 2.98 & \underline{84.15} & 54.35 & \textbf{84.39} \\
&

& Generalization   & 0.00 & 0.00 & 0.00 & 0.00 & \underline{44.10} & 0.36 & \textbf{50.18}  \\
                             &                                           
& Locality   & 100.0 & 70.66 & 89.28 & 90.86 & \underline{91.18} & \textbf{99.75} & 88.04 \\
                             &                                           
& Portability   & 27.04 & 3.19 & 26.93 & 33.43 & \underline{65.84} & 27.04 & \textbf{69.43} \\ 

\midrule
\multirow{7}{*}{\textbf{\wikirecent}} & \multirow{3}{*}{GPT2-XL}                    
& Reliability   & 2.61 & 6.48 & 11.53 & 49.37 & 85.62 & \textbf{99.68} & \underline{85.70} \\
&

& Generalization   & 1.58 & 2.21 & 5.37 & 10.98 & \textbf{52.76} & 1.58 & \underline{51.66}  \\
                             &                                           
& Locality   & 100.0 & 26.58 & 83.87 & \underline{87.12} & 57.94 & \textbf{100.0} & 64.61 \\
                             &                                           
& Portability   & 17.19 & 16.78 & 10.31 & 30.41 & \underline{42.26} & 17.73 & \textbf{47.36} \\ 
\cmidrule{2-10} 
                             & \multicolumn{1}{c}{\multirow{3}{*}{LLaMA-2}} 
& Reliability   & 9.87 & 6.16 & 9.79 & 15.88 & 82.31  & \underline{83.72} & \textbf{83.73} \\
&

& Generalization   & 7.27  & 3.87 & 6.64 & 0.08 & \underline{54.66} & 7.35 & \textbf{55.92}  \\
                             &                                           
& Locality   & 100.0 & 70.66 & \underline{89.28} & 88.88 & 78.57 & \textbf{99.98} & 87.04 \\
                             &                                           
& Portability   & 43.52 & 3.15 & 43.26 & 43.52 & \underline{60.84} & 44.13 & \textbf{62.39} \\ 

\midrule

\multirow{3}{*}{\textbf{\convsent}} & \multirow{3}{*}{GPT2-XL}                    
& Reliability   & 40.74 & 7.48 & 37.47 & 53.07 & \underline{54.67}  & 40.74 & \textbf{65.43} \\

                             &                                           
& Locality   & 100.0 & 42.86 & 87.47 & 94.58 & \underline{96.58} & \textbf{100.0}  & 94.27 \\    
                            & 
& Fluency   & 613.13 & 548.55 & 396.43 & \underline{615.61} & 601.93 & 414.03 & \textbf{617.65} \\          


\midrule

\multicolumn{10}{c}{\textbf{\small Hold Out Editing}} \\

\midrule

\multirow{8}{*}{\textbf{\zsre}} & \multirow{4}{*}{GPT2-XL}                    
& Reliability   & 0.00 & 0.11 & 0.00 & 13.50 & \underline{40.79}  & 0.00 & \textbf{82.83} \\
&

& Generalization   & 0.00 & 0.08 & 0.10 & 10.13 & \underline{31.15}  & 0.00 & \textbf{78.40}  \\
                             &                                           
& Locality   & 100.0 & 74.06 & \underline{95.66} & 82.59 & 94.79 & \textbf{100.0} & 94.57 \\
                             &                                           
& Portability   & 47.07 & 0.96 & 0.39 & 43.90 & \underline{45.08} & \textbf{47.07} & 40.84 \\ 
\cmidrule{2-10} 
                             & \multicolumn{1}{c}{\multirow{4}{*}{LLaMA-2}} 
& Reliability   & 0.00 & 2.23 & 0.00 & 2.70 & \textbf{76.95}  & 0.00 & \underline{76.57} \\
&

& Generalization   & 0.00 & 1.93 & 0.00 & 0.19 & \underline{67.89}  & 0.00 & \textbf{70.11}  \\
                             &                                           
& Locality   & 100.0 & 98.89 & \underline{99.66} & 95.15 & 90.14 & \textbf{100.0} & 94.16 \\
                             &                                           
& Portability   & 56.66 & 0.54 & 0.87 & 48.02 & \textbf{58.63} & 56.66 & \underline{58.19} \\ 
\bottomrule
\end{tabular}
\end{adjustbox}
}
}
\caption{
\textbf{Multi-Task Editing Setting}: Editors train on a hybrid of CounterFact, Recent, and ConvSent datasets, and test on their specific test sets.
\textbf{Hold Out Editing Setting}: The abovementioned editors are tested on ZsRE (OOD data).
All metrics are ``the higher, the better''.}
\label{tab:main-results}  
\end{table*}

\section{Experiments}
\subsection{Experimental Settings}
\paragraph{Editing Models.}
We conduct experiments on GPT2-XL(1.5B)~\cite{GPT-2} and LLaMA-2-Base (7B)~\cite{LLaMA}.

\paragraph{Baselines.}
In this paper, we compare our method with \texttt{FT-L} method, which involves fine-tuning a specific layer's FFN identified via causal tracing in ROME~\cite{ROME}.
We further compare our method with preserve models’ parameters editing baselines including \texttt{CaliNet} and \texttt{GRACE}, and modify models’ parameters editing baselines including \texttt{MEND} and \texttt{KE}.



\subsection{Metrics}
We apply several evaluation metrics consistent with those described in~\cite{yao2023editing}.


\paragraph{Reliability.} 
Reliable editing is defined when the post-edit model $f_{\theta^{\prime}}$ generates the target answer correctly for the case $(x_i,y_i)$. 
Reliability is assessed based on the average accuracy of the edit case. 
\begin{equation}
\mathbb{E}_{x_{\mathrm{i}}^{\prime}, y_{\mathrm{i}}^{\prime} \sim \left\{\left(x_{\mathrm{i}}, y_{\mathrm{i}}\right)\right\}} \mathbb{1} \left\{\operatorname{argmax}_y f_{\theta^{\prime}}\left(y \mid x_{\mathrm{i}}^{\prime}\right)=y_{\mathrm{i}}^{\prime}\right\}
\end{equation}

\paragraph{Generalization.} The post-edit model $f_{\theta^{\prime}}$ should predict the equivalent neighbor $N(x_{\mathrm{i}}, y_{\mathrm{i}})$, like rephrased sentences, and its performance is assessed by the average accuracy on examples uniformly sampled from this equivalence neighborhood\footnote{We follow \cite{KnowledgeEditor,yao2023editing} to evaluate the rephrase-based generalization.}.
\begin{equation}
\mathbb{E}_{x_{\mathrm{i}}^{\prime}, y_{\mathrm{i}}^{\prime} \sim N\left(x_{\mathrm{i}}, y_{\mathrm{i}}\right)} \mathbb {1} \left\{\operatorname{argmax}_yf_{\theta^{\prime}}\left(y \mid x_{\mathrm{i}}^{\prime}\right)=y_{\mathrm{i}}^{\prime}\right\}
\end{equation} 

\paragraph{Locality.} Editing should be implemented locally, ensuring that the post-edit model $f_{\theta^{\prime}}$ preserves the outputs for out-of-scope examples $O(x_i,y_i)$.
Therefore, locality is measured by the rate at which $f_{\theta^{\prime}}$ maintains the same predictions as the pre-edit model $f_\theta$.
\begin{equation}
\mathbb{E}_{x_{\mathrm{i}}^{\prime}, y_{\mathrm{i}}^{\prime} \sim O\left(x_{\mathrm{i}}, y_{\mathrm{i}}\right)} \mathbb {1} \left\{f_{\theta^{\prime}}\left(y \mid x_{\mathrm{i}}^{\prime}\right)=f_{\theta}\left(y \mid x_{\mathrm{i}}^{\prime}\right) \right\}
\end{equation}

\paragraph{Portability.} Portability, proposed by \cite{yao2023editing}, gauges the edited knowledge application of the post-edit model $f_{\theta^{\prime}}$. \cite{yao2023editing} adds $P(x_i, y_i)$ to the existing dataset and calculates Portability by the edited model's average accuracy on these new reasoning parts.
\begin{equation}
\mathbb{E}_{x_{\mathrm{i}}^{\prime}, y_{\mathrm{i}}^{\prime} \sim P\left(x_{\mathrm{i}}, y_{\mathrm{i}}\right)} \mathbb {1} \left\{\operatorname{argmax}_y f_{\theta^{\prime}}\left(y \mid x_{\mathrm{i}}^{\prime}\right)=y_{\mathrm{i}}^{\prime}\right\}
\end{equation}

\paragraph{Fluency.} Fluency measures the edited model $f_{\theta^{\prime}}$'s generative performance by using a weighted average of bi-gram and tri-gram entropies to evaluate text diversity. Lower values suggest higher repetition.

\begin{equation}
\text{Fluency} = \frac{\sum w_n \cdot H_n}{\sum w_n}
\end{equation}
where $H_n$ represents n-grams entropy (bi-gram for $n=2$, tri-gram for $n=3$) and $w_n$ the respective weights.
This metric is specifically tailored for \text{\convsent} testing, where longer responses require scrutiny of the model's fluency.

\subsection{Main Results}
We evaluate the efficacy of \textsc{InstructEdit} by examining three key aspects: \textbf{Multi-Task Editing}, \textbf{Hold Out Editing}, and \textbf{Transfer to Unseen Instruction}.

\paragraph{Multi-Task Editing Results.}
Table~\ref{tab:main-results} presents the corresponding results.
\texttt{FT-L}~\cite{yao2023editing} exhibit subpar performance in Reliability for multi-task editing, which we believe is due to the interference of the original models' prior knowledge, complicating the editing process.
Moreover, we notice that \texttt{FT-L} does not enhance Portability or Generalization, as expected due to its focus on fitting updated knowledge.
Our experiments reveal that \texttt{FT-L} substantially reduces the original model's parameter knowledge, significantly lowering Locality.
Preserve Models’ Parameters Editing Methods like \texttt{CaliNet}~\cite{CALINET} maintain backbone model integrity, resulting in high Stability, but their performance in other metrics is unsatisfactory.
Similar to \texttt{FT-L}, \texttt{CaliNet} overfits updated knowledge, leading to poor Generalization and Portability, but it has better Locality than \texttt{FT-L} as it doesn't alter the original parameters of the LLMs.
While \texttt{GRACE} represents the state-of-the-art of Preserve Models’ Parameters Editing Methods, delivering outstanding Reliability and Locality, it falls short in the metrics of Generalization and Portability.
Modify Models' Parameters Editing Methods, such as \texttt{KE}~\cite{KnowledgeEditor} and \texttt{MEND}~\cite{MEND}, surpass previous editing approaches in effectiveness. Both \texttt{MEND} and \texttt{KE} excel across all metrics, achieving a balance between Reliability and Locality. 
This is attributed to their optimization objectives that limit update extents, thus enabling editors to adjust parameters while preserving model stability.
We can observe our \textsc{InstructEdit} improves editing precision and control with instruction-guided methods, reaching effectiveness akin to advanced hypernets like MEND and KE.
While MEND and KE yield effective editing results, their performance is suboptimal on OOD data, with editing in In-Distribution data often causing misdirection in the update trajectory of the posterior vector space.
However, we find that providing specific command hints to the Editor can substantially alleviate this issue.

\paragraph{Hold Out Editing Results.}
To evaluate the adaptability of knowledge editing methods to OOD data, we devise the ``Hold Out Editing Setting''.
In this setup, the editor is trained using datasets like \wikirecent, \wikicf, and \convsent, and then evaluated on \zsre. 
From Table~\ref{tab:main-results}, we notice a linear decline in the performance of all previous knowledge editing baselines when applied to OOD data.
This decline can be attributed primarily to the editor's limitations in defining new editing tasks and its insufficient generalization capability for handling OOD scenarios.
We observe that \textsc{InstructEdit} can effectively address these challenges.
Note that such robust generalization abilities are mainly inherent in instruction tuning, a synergy that enables \textsc{InstructEdit} to attain performance levels on par with single-task editing, even on task datasets that are unseen during the training phase.

\begin{figure}[ht]
\centering 
\includegraphics[width=0.47\textwidth]{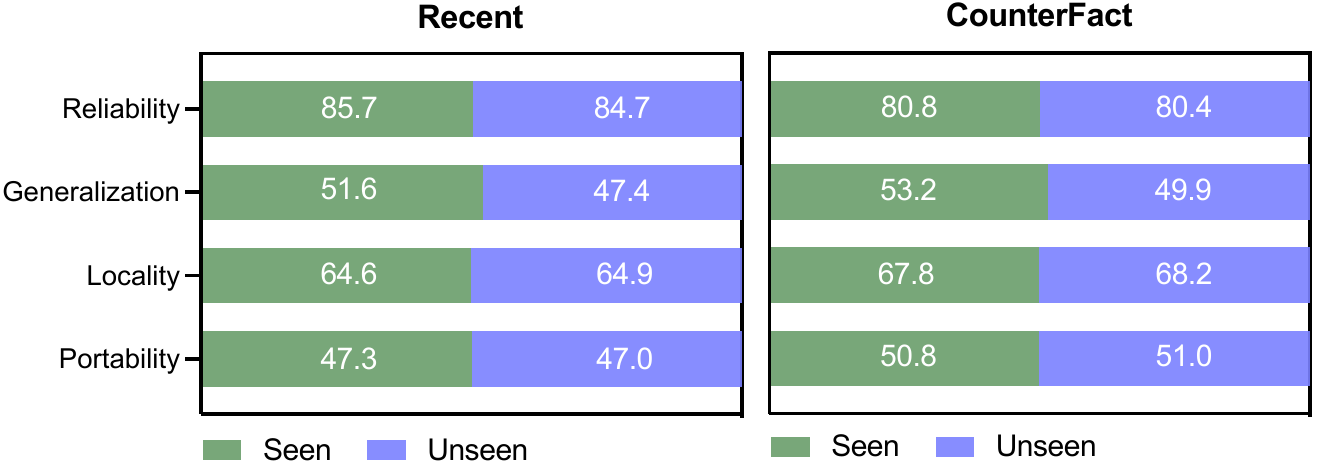}
\caption{
\textsc{InstructEdit} demonstrates proficiency in generalizing to Unseen instructions (unseen instructions introduced in Section \ref{sec:instructedit}), achieving results comparable to Seen instructions.
}
\label{fig:instruct_generality}
\end{figure}

\paragraph{Transfer to Unseen Instructions.}
To delve deeper into the generalizability of Instruction-based Editing, we evaluate \textsc{InstructEdit}'s capacity with instructions that have not been encountered previously.
This setting is different from the hold-out editing setting since we still use the data in \wikicf, \wikirecent, \convsent, and \zsre, but with new instructions. 
Specifically, as outlined in Section~\ref{sec:instructedit}, we construct five novel, unseen instructions to assess the Editor's proficiency in generalizing instructions. 
Observations from Figure~\ref{fig:instruct_generality} reveal that the Editor is indeed capable of adapting to these Unseen Instructions. 
It is noteworthy that utilizing instructions on which the Editor has been trained can result in enhanced editing performance.
Thus, \textsc{InstructEdit} can achieve comparable outcomes by employing instructions that are semantically akin to those encountered during training.
These empirical results also indicate that we can develop an Editor to follow human instructions and we leave this for future works.

\begin{figure*}[t]
\centering 
\includegraphics[scale=0.6]{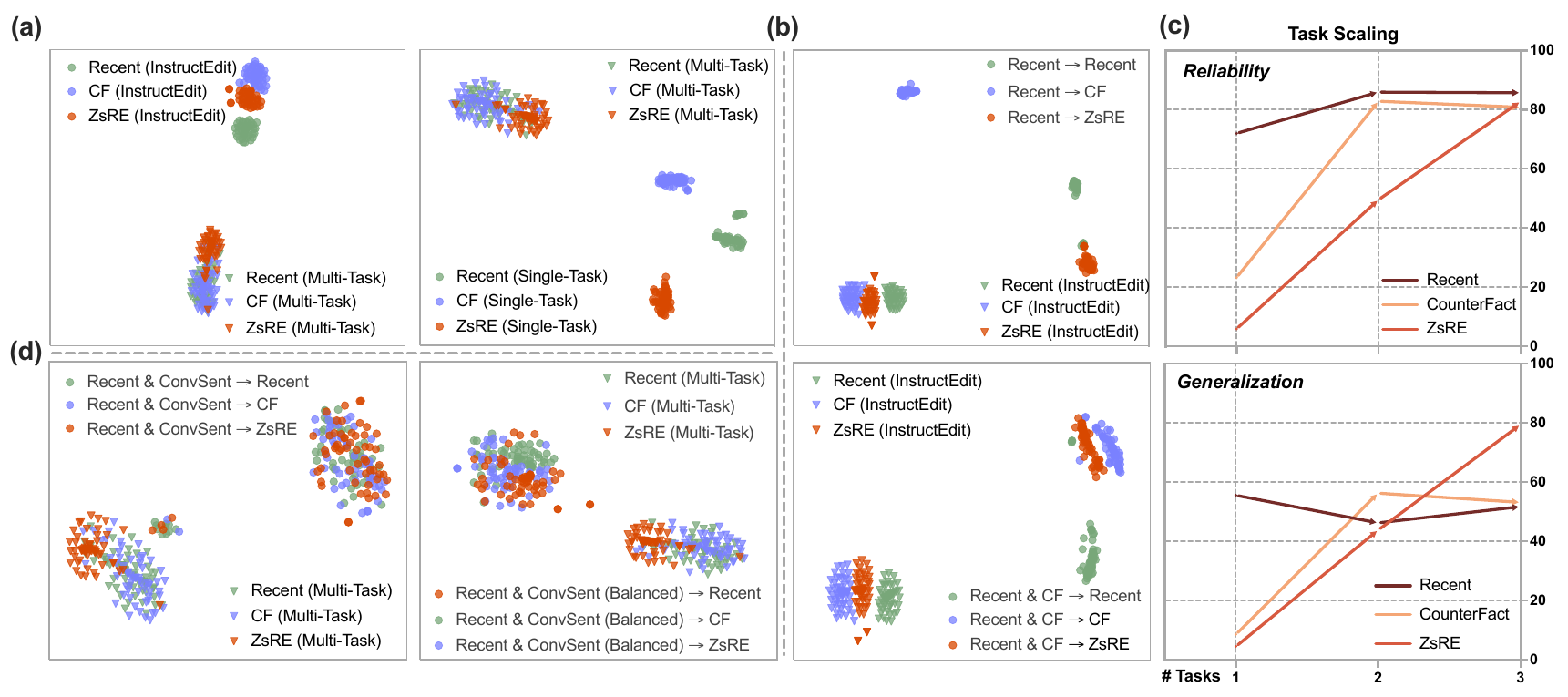}
\caption{
(a) Compares instruction effects on knowledge editing gradient $\tilde{\nabla}_{u_\ell}$. \textbf{Recent (InstructEdit)} and \textbf{Recent (Multi-Task)} illustrate $\tilde{\nabla}_{u_\ell}$ on \text{\wikirecent} using \textsc{InstructEdit} and MEND in multi-task settings, respectively. \textbf{Recent (Single-Task)} shows MEND's results of training on \text{\wikirecent} alone. (b) Demonstrates task scaling's impact on \textsc{InstructEdit}, with \textbf{Recent$\rightarrow$ZsRE} for training on \text{\wikirecent} and testing on \text{\zsre}, and \textbf{Recent\&CF$\rightarrow$ZsRE} for joint training on \text{\wikirecent}, \text{\wikicf}, and testing on \text{\zsre}. (c) Illustrates the reliability and generalization performance across task scaling. (d) Balances \text{\convsent} by extracting 1,427 entries for \textbf{ConvSent (Balanced)}.}
\label{fig:task_num}
\end{figure*}

\subsection{Why Instruction Helps Multi-Task Editing?}
We analyze the principal components of the editing area $\vec{\nabla}_{u_\ell}$ using t-SNE, as presented in Section \ref{sec:instructedit}, which is generated by the editor for specific layers of LLMs. 
Our underlying assumption is that these principal components encapsulate the intrinsic characteristics of the editing area involved in editing the data.
Specifically, we focus our analysis on cases where the conventional editing methods fall short, while \textsc{InstructEdit} demonstrates effectiveness.

\paragraph{Finding 1: Instruction can Help Control Optimization Direction.} 
As observed in Table \ref{fig:multitask_editing}, MEND exhibits subpar performance in multi-task scenarios, particularly in terms of Reliability and Generalization, where it is significantly outperformed by \textsc{InstructEdit}. 
Upon analyzing the left panel in Figure \ref{fig:task_num}(a), we observe that MEND, when handling multi-task editing, tends to show significant overlap in editing area across different tasks. 
This overlap could potentially cause MEND to not only confuse previously learned tasks but also struggle in effectively generalizing to new tasks with shifted distribution compared to the training tasks. 
However, by introducing instructions, \textsc{InstructEdit} can effectively control the knowledge editing gradient and encourage distinct separation with adequate margin in the editing area for various tasks, which aligns with the distribution observed in the single-task training setting in the right panel of Figure \ref{fig:task_num}(a).
Furthermore, the discriminative editing area in \textsc{InstructEdit} is adaptable to OOD data, which leads to superior knowledge editing when handling new tasks, while maintaining performance comparable to models trained on single tasks on ID tasks.

\paragraph{Finding 2: More Tasks, Stronger OOD Generalization.} Figure \ref{fig:task_num}(b) illustrates that when \textsc{InstructEdit} is trained on a single task, the editing areas of the three tasks appear somewhat discriminative. 
Instead, the performance of the corresponding tasks is suboptimal, as demonstrated in Figure \ref{fig:task_num}(c).
We think that even though instructions aid in distinguishing different tasks, the knowledge learned from a single task struggles to generalize to others. 
By scaling the number of tasks in training, we notice that the editing areas of \textsc{InstructEdit} for various tasks almost see no overlap, and editing reliability improves correspondingly in Figure \ref{fig:task_num}(c). 
Furthermore, as the scope of tasks broadens, the directions of knowledge editing gradient of different tasks start to converge, yet they retain their relative margin. 
Intuitively, \textsc{InstructEdit} trained across diverse domains harnesses these domain-related instructions to extrapolate effectively to new, unseen domains, while offering a trade-off between specialized knowledge adaptation and broad generalization. Nevertheless, it is crucial to acknowledge that a scalability bottleneck might be encountered, and confronting entirely new types of editing tasks, such as cross-linguistic tasks, will introduce further complexities.

\paragraph{Finding 3: Improving Performance with Appropriate Data Proportion.} 
 In preliminary experiments, we notice task imbalances impede proper multi-task training and cause a significant performance decline when \text{\convsent} is involved in the training. 
Hence, we contemplate balancing the data proportions across different tasks. 
By observing Figure \ref{fig:task_num}(d), we find that the knowledge editing gradient directions become more regular after data balancing and editing reliability of the editor increases from 18.23 to 25.55 on the OOD tasks.
Additionally, we find that task imbalances lead to the editor inducing editing gradients of relatively large magnitudes, and the gradient magnitude distributions for each task vary significantly, which appears to be a key factor influencing the editor's generalization.
This result confirms the significance of appropriate data proportions for multi-task editing.

\section{Discussion and Conclusion}

We focus on a new problem of knowledge editing for LLMs: generalizing to new tasks.
We introduce multi-task editing, illustrating the limitations of existing knowledge editing approaches in task transferability and presenting a viable solution \textsc{InstructEdit}.
The proposed approach can effectively guide the Editor for precise editing, with its effectiveness confirmed through comprehensive experiments and visualization analysis.





\section*{Acknowledgments}

We would like to express gratitude to the anonymous reviewers for kind comments, EasyEdit \cite{easyedit} and many other related works for their open-source contributions as well as GPT-4 Service \cite{DBLP:journals/corr/abs-2303-08774}. 
This work was supported by the National Natural Science Foundation of China (No. 62206246), the Fundamental Research Funds for the Central Universities (226-2023-00138), Zhejiang Provincial Natural Science Foundation of China (No. LGG22F030011), Yongjiang Talent Introduction Programme (2021A-156-G), CCF-Tencent Rhino-Bird Open Research Fund, and Information Technology Center and State Key Lab of CAD\&CG, Zhejiang University.


\bibliographystyle{named}
\bibliography{ijcai24}


\end{document}